\documentclass[letterpaper, 10 pt, conference]{ieeeconf}  

\IEEEoverridecommandlockouts                              

\usepackage[%
  style=numeric,
  backend=biber,
  natbib=true
]{biblatex}

\usepackage{amsmath,amssymb}
\usepackage{caption}
\usepackage{subcaption}
\usepackage{graphicx} 

\title{\LARGE \bf
Acceleration based PSO for Multi-UAV Source-Seeking
}

\author{Adithya Shankar$^{1}$, Harikumar Kandath$^{1}$ and J. Senthilnath$^{2}$
\thanks{}
\thanks{$^{1}$ Robotics Research Center, International Institute of Information Technology, Hyderabad-500032, India. 
        {E-mails: $adithyashankar.b@gmail.com$, $harikumar.k@iiit.ac.in$}}%
\thanks{$^{2}$Institute for Infocomm Research, Agency for Science, Technology and Research (A*STAR), Singapore, 138632. E-mail: $J\_Senthilnath@i2r.a$-$star.edu.sg$}
}

\DeclareUnicodeCharacter{2212}{-} 
\begin{document}

\maketitle
\thispagestyle{empty}
\pagestyle{empty}

\begin{abstract}

This paper presents a novel algorithm for a swarm of unmanned aerial vehicles (UAVs) to search for an unknown source. The proposed method is inspired by the well-known PSO algorithm and is called acceleration-based particle swarm optimization (APSO) to address the source-seeking problem with no a priori information. Unlike the conventional PSO algorithm, where the particle velocity is updated based on the self-cognition and social-cognition information, here the update is performed on the particle acceleration. A theoretical analysis is provided, showing the stability and convergence of the proposed APSO algorithm. Conditions on the parameters of the resulting third order  update equations are obtained using Jury’s stability test. High fidelity simulations performed in CoppeliaSim \footnote{https://www.coppeliarobotics.com/}, shows the improved performance of the proposed APSO algorithm for searching an unknown source when compared with the state-of-the-art particle swarm-based source seeking algorithms. From the obtained results, it is observed that the proposed method performs better than the existing methods under scenarios like different inter-UAV communication network topologies, varying number of UAVs in the swarm, different sizes of search region, restricted source movement and  in the presence of measurements  noise.
\end{abstract}
\section{INTRODUCTION}

The objective of source-seeking problem in robotics is to locate a stationary or moving source by measuring the signal strength that is emanated from the source. In large and unknown outdoor environments, unmanned aerial vehicles (UAVs) becomes the preferred robotic platform for such applications.  Deploying multiple UAVs for such a mission proves to be efficient than using a single UAV [6]. However, source seeking based on multiple UAVs  has its own challenges, like to develop an efficient cooperative framework among UAVs to complete the task. Stochastic search methods became a natural choice when there is no a priori information available on the probability distribution of the source location [5].   Swarm intelligence based algorithms, such as Particle Swarm Optimization (PSO) [13], and Fruit Fly Optimization Algorithm (FOA) [9],  are widely used for solving the source-seeking problem. This study focuses on the PSO algorithm and its variants for solving the source-seeking problem.

Particle swarm optimization (PSO) is a swarm-based algorithm developed by Kennedy et al. [8], which is based on the collective social behavior of the particles. The method performs well and is capable of optimizing the source-seeking problem. Over the last decade, a lot of the research has focused on using PSO-based algorithms for the source-seeking problem. This research has led to many variants of PSO such as the Darwinian PSO (DPSO) [11] algorithm to overcome the local optima problem of PSO. The algorithm does this by initializing multiple swarms that compete with each other in order to simulate a phenomenon that is similar to natural selection. PSO uses optimal computing budget allocation to improve the performance in the presence of noise for multiple UAVs performing required task [4]. This study focuses on modified PSO to locate the source in the presence of noise. 

Two other variations of PSO that have gained popularity are Robotic PSO (RPSO) and Robotic Darwinian PSO (RDPSO) [1]. RPSO consists of a population of robots that collectively searches source in the search space. RDPSO is an extension of RPSO that introduces a “punish-reward” mechanism to reduce the shared information between the UAVs since multiple swarms are formed. The main goal behind this technique is to prevent robots from erroneously converging towards local optima.  In both of these algorithms, the team of robots is capable of performing the source-seeking task in a distributed manner. The other  popular PSO based source seeking algorithm is the standard-PSO (SPSO) [14]. The SPSO implements the optimized hyperparametrs for the multi-robot source seeking problem and proves to  be better than the conventional PSO based algorithm [14]. It also evaluates the performance of the algorithm under three different inter-robot communication network topologies. In the above-mentioned PSO variants, the hyperparameters are kept constant during the entire robotic search mission. Whereas, the hyperparameters are adaptively updated in Adaptive-Robotic PSO algorithm [2].
In [2], the authors showed that ARPSO performed better than RPSO and RDPSO. Hence, in this paper we use ARPSO and SPSO for the comparative study with the proposed APSO algorithm.

This paper presents a novel update rule for the particle's position. The original PSO can be portrayed as a second-order system with the particle's position being dependent on the particle's velocity. In our proposed method, the  particle position is updated based on it's velocity and acceleration resulting in a third order dynamical system.  We term this new algorithm as "APSO" (Acceleration Based PSO).

The major contributions of this paper are the following.
\begin{itemize}
    \item A new PSO based source seeking algorithm is presented that utilizes a third order update rule for particle's position.  The acceleration of the particle is updated using the self-cognition and social-cognition component unlike the  existing PSO based methods where the particle's velocity is updated through the self-cognition and social-cognition component.
    \item Detailed convergence and stability  analysis of the proposed algorithm  is provided.
    \item The condition on the values to be taken by the hyperparameters of the algorithm to ensure stability is established through Jury's stability test.
    \item A detailed performance comparison is provided with the state-of-the-art PSO based source seeking algorithms, showing the improved performance of the proposed method quantified in terms of average source seeking time, average number of iterations taken and the average swarm distance. A high fidelity simulation is executed in CoppeliaSim for the performance comparison. The improvement in performance of the proposed algorithm is shown to be consistent across  scenarios like different inter-UAV communication network topologies, varying number of UAVs in the swarm, different sizes of search region, restricted source movement and  in the presence of measurements  noise.
\end{itemize}

The paper is structured in the following manner. Section 2 briefly describes PSO, SPSO, and ARPSO. Section 3 details the  APSO algorithm along with its convergence and stability analysis. The supporting results and discussions are presented in section 4.  In section 5,  the paper is concluded with a mention of the future work.

\section{Background}
This section provides the details on the UAV motion, details of the sensor carried by the UAV,  and also a description of the existing PSO based multi-robot source seeking algorithms.

\subsection{UAV and sensor  description}
The swarm of $n$ UAVs are randomly initialized at different locations in the search space to locate the source position solely based on the sensor information. The position of the UAVs are denoted with respect to an inertial frame of reference as $X_I-Y_I-Z_I$. All the UAVs are assumed to be flying at a fixed altitude different from each other to avoid inter-UAV collision, and hence the $Z_I$  coordinate is not discussed here. The position coordinate of the $i^{th}$ UAV at any given time $t$ is denoted by  $P_i(t)=[x_i(t),\,y_i(t)]^T$.  The UAVs fly at a constant speed between the successive waypoints.

The source is assumed to be  located with in the search space and emits a signal having a power that  exponentially decays with the distance  from it. Each UAV carries a sensor that measures the signal strength emitted by the source using the below model.

\begin{equation}
    S_{mi} = S_{s}e^{{-\alpha d_{i}^{2}}} + \delta S
    \label{noise1}
\end{equation}
where $S_{mi}$ is the power measured by the $i^{th}$ UAV located at a distance $d_i$ from the source, $S_s$ is the source power and $\alpha>0$ is a constant. The measurement noise is denoted by $\delta S$ and is random with a magnitude proportional to the measured power $S_{mi}$. The discrete time version of the position coordinates is used in the subsequent sections and is denoted by $P_i(k)=[x_i(k),\,y_i(k)]^T$, where $t=k\tau$, $k$ is the discrete time instant and $\tau$ is the sampling time in seconds.

\subsection{Algorithms for PSO based source-seeking}
A brief overview of the popular PSO based source-seeking algorithms is presented below. Equations are provided for a single-axis $X_I$ and analogous equations follows for the $Y_I$ axis also.\\
a) \textit{Regular PSO}: The  conventional PSO used for optimization is implemented for source-seeking problem in [3]. The velocity ($v_i(k)$) and the position ($x_i(k)$) updates for the $i^{th}$ robot for $k^{th}$ sampling time instant is given by,

\begin{equation}
\begin{split}
    v_i(k+1) = v_{i}(k) + R(0, c_{1})(x_{ib}(k) − x_{i}(k))\\ + R(0, c_{2})(x_{gb}(k) − x_{i}(k)) 
    \end{split}
    \label{eq1}
\end{equation}

\begin{equation}
  x_{i}(k+1) = x_{i}(k) + v_i(k+1)T 
   \label{eq2}
\end{equation}

Here  $T$ is a scaling factor (typically $T$=1 is used), $c_1>0$, $c_2>0$, $R(a,\,b)$ denotes a random number from uniform distribution within the interval $[a,\,b]$,   $x_{ib}(k)$ is the best individual position and $x_{gb}(k)$ is the best global position.

b) \textit{Standard Particle Swarm Optimization (SPSO)}: SPSO is an extension to the regular PSO algorithm [14] with an inertia factor added in the velocity update of the $i^{th}$ particle is given by,

\begin{equation}
\begin{split}
    v_i(k+1) = \omega v_{i}(k) + R(0, c_{1})(x_{ib}(k) − x_{i}(k))\\ + R(0, c_{2})(x_{lb}(k) − x_{i}(k)) 
    \end{split}
     \label{eq3}
\end{equation}
The parameter values used are $\omega$=0.721 and $c_1=c_2=1.193$ to achieve the best performance. Another important difference from the regular PSO is the incorporation of communication constraints among the particles. The global best position $x_{gb}(k)$ is replaced by the local best $x_{lb}(k)$ considering the particles that can communicate with each other.\\
c) \textit{Adaptive-Robotic PSO (ARPSO)}: The ARPSO algorithm prevents the swarm from converging at a local minima. The algorithm is also capable of handling obstacles that are present in the environment [2]. 
\begin{equation}
\begin{split}
    v_i(k+1) = \omega_{i}v_{i}(k) + R(0, c_{1})(x_{ib}(k) − x_{i}(k))\\ + R(0, c_{2})(x_{gb}(k) − x_{i}(k))+ R(0, c_{3})(x_{a}− x_{i}(k))
    \end{split} 
     \label{eq4}
\end{equation}
where $x_{a}$ is an attractive position that is located away from the obstacle. The parameter  $c_{3}$ is set to 0 for obstacle free environment. Unlike in SPSO, the parameter $\omega_{i}$ in ARPSO is not constant. The value of $\omega_{i}$ is different for different UAVs and  also changes for each iteration. 

\section{Acceleration-based PSO (APSO) for source-seeking}
In the proposed APSO, the particle acceleration is also updated apart from velocity as in the above-mentioned PSO and its variants to generate a new waypoint. This additional equation transforms the input to output relation to a third-order system. The following equations from (\ref{eq5}) to (\ref{eq7}) shows the waypoint update equations for the $i^{th}$ UAV.

\begin{equation}
\begin{split}
    a_i(k+1) = w_{1}a_i(k) + R(0, c_1)(x_{ib}(k) − x_{i}(k)) \\+ R(0, c_2)(x_{gb}(k) − x_{i}(k)) 
    \end{split}
     \label{eq5}
\end{equation}

\begin{equation}
   v_{i}(k+1) = w_{2}v_{i}(k) + a_i(k+1)T  
    \label{eq6}
\end{equation}

\begin{equation}
  x_{i}(k+1) = x_{i}(k) + v_i(k+1)T 
   \label{eq7}
\end{equation}
where $a_i(k)$ denotes the particle acceleration for the $i^{th}$ UAV. 
\subsection{Convergence Analysis}
By simplifying (\ref{eq5}) to (\ref{eq7}), the following relation is obtained with $r_1=R(0, c_1)$ and $r_2=R(0, c_2)$.

\begin{equation}
\begin{split}
  x_{i}(k+1) =(1+w_1+w_2-r_1T-r_2T)x_{i}(k)\\+(-w_1-w_2-w_1w_2)x_{i}(k-1)+w_1w_2 x_{i}(k-2)\\+T(r_1x_{ib}(k)+r_2x_{gb}(k))
  \end{split}
  \label{eq8}
\end{equation}

A steady state condition  for $x_{i}(k)$  (denoted by $x_{i}(ss)$) in (\ref{eq8}) is achieved when i) $r_1$ and $r_2$  are  made constant or when ii) $x_{ib}(ss)=x_{gb}(ss)$. When $r_1$ and $r_2$  are  made constant  (denoted by $r_{1c}$ and $r_{2c}$ respectively), the following result is obtained.
\begin{equation}
  x_{i}(ss)=   \lim_{k\to\infty} x_{i}(k) = \frac{r_{1c}x_{ib}(ss)+r_{2c}x_{gb}(ss)}{r_{1c}+r_{2c}}
     \label{eq9}
\end{equation}
When $x_{ib}(ss)=x_{gb}(ss)$, 
\begin{equation}
     x_{i}(ss)= x_{ib}(ss)=x_{gb}(ss)
     \label{eq10}
\end{equation}

\subsection{Stability Analysis}
The equation describing the evolution of $x_i(k)$ given in  (\ref{eq8}) represents a third order dynamical system. Applying $z$-transform on (\ref{eq8}) yields the following transfer function.
\begin{equation}
    G(z)=\frac{X(z)}{U(z)}=\frac{z^{-1}}{1+a_1z^{-1}+a_2z^{-2}+a_3z^{-3}}
    \label{eq11}
\end{equation}
where the coefficients $a_1=-1-w_1-w_2+r_1T+r_2T$, $a_2=w_1+w_2+w_1w_2$, $a_3=-w_1w_2$ and the input 
$U(z)=T(r_1X_{ib}(z)+r_2X_{gb}(z))$. The stability of $G(z)$ depends upon the roots of the polynomial given below.
\begin{equation}
    H(z)=1+a_1z^{-1}+a_2z^{-2}+a_3z^{-3}
    \label{eq12}
\end{equation}
Since $r_1$ and $r_2$ are random numbers varying between $[0,\,c_1]$ and $[0,\,c_2]$ respectively, the coefficient $a_1 \in [-1-w_1-w_2,\,-1-w_1-w_2+T(c_1+c_2)]$. Since $H(z)$ is a third order uncertain polynomial, it is enough to confirm that the extreme polynomials
$H(z)$ at $a_1=-1-w_1-w_2$ (the coefficient and corresponding polynomial denoted by $a_{1l}$ and $H(z,a_{1l})$ respectively)  and  $H(z)$ at $a_1=-1-w_1-w_2+T(c_1+c_2)$ (the coefficient and corresponding polynomial denoted by $a_{1u}$ and $H(z,a_{1u})$ respectively) does not have roots outside the unit circle [7]. Applying the Jury's stability criterion for $H(z,a_{1l})$ and $H(z,a_{1u})$, the following conditions are obtained on the parameters $w_1$, $w_2$, $T$, $c_1$  and $c_2$.
\begin{equation}
     \frac{2}{T}(1 + w_{1} + w_{2} + w_{1}w_{2})>c_1+c_2
     \label{c1}
\end{equation}
\begin{equation}
    |w_{1}w_{2}| < 1 
    \label{c2}
\end{equation}
\begin{equation}
    |(1-w_{1}w_{2})(w_{1} + w_{2}) + (w_{1}w_{2})(c_1T+c_2T)| < |1-(w_{1}w_{2})^2|
    \label{c3}
\end{equation}
\begin{equation}
    |w_{1} + w_{2}|  < |1+w_{1}w_{2}|
    \label{c4}
\end{equation}



\section{Results and Discussion}
\begin{figure}[t]
\includegraphics[width = 8 cm]{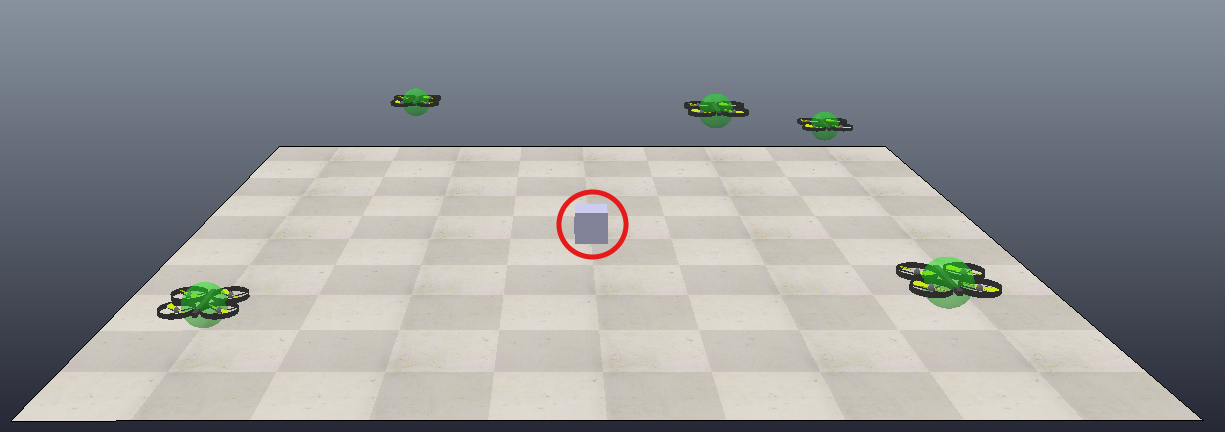}
\caption{Search scenario showing 5 UAVs looking for the source (shown with in a red circle). Note that each UAV maintain an altitude different from the others to avoid inter-UAV collision. }
\label{sspace}
\end{figure}

In this section, the performance of APSO is compared against that of ARPSO and the standard PSO (SPSO). The obtained results are verified by simulation performed in the CoppeliaSim simulator (\url{www.youtube.com/watch?v=ad_GD3NQR-w}). Coppeliasim (formerly known as ”V-REP”) is high fidelity and scalable robotic simulator used for testing and verification of different algorithms before deployment in the actual hardware [10]. All the three search algorithms (APSO, ARPSO, and SPSO) were implemented using python 3.7 in a conda based environment. A typical search scenario with 5 UAVs  is shown in Fig. \ref{sspace}. The search space considered here is in the shape of a square.  For each simulation, the UAVs starts from randomly initialized locations along the boundary of the search space. All the UAVs move from a waypoint to the next with a constant speed of 10 $m/s$. For all the simulations, the source is considered to be at the center of the search space. The algorithm is terminated when any of the UAV reaches a distance of 0.1 $m$ from the source, in the case of measurement noise this value is increased to 0.5 $m$.

In the implementation of APSO, the values of $w_{1}$ and $w_{2}$ are taken as 0.675 and -0.285 respectively. Both of the parameters $c_{1}$ and $c_{2}$ are selected as 1.193, and the value of T is taken as 1.  It can be  verified that the above-mentioned parameter setting satisfies the stability conditions mentioned in (\ref{c1}) to  (\ref{c4}). The values of the parameters of SPSO and ARPSO algorithm are selected from [14] and [2] respectively.
 






\subsection{Metrices for the performance comparison of  source-seeking algorithms}
The metrics used for performance comparison are  given below. The metrices are averaged over 1000 Monte Carlo runs.

\begin{itemize}
\item \textbf{Average source seeking time ($\mu(T_s)$)}: This metric quantifies the time efficiency of the underlying source seeking algorithm and is defined in (\ref{m1}). Source seeking time is taken as the time duration from the start of the search  till the source has been located by one or more members of the UAV swarm.  

\begin{equation}
     \mu(T_s) = \frac{\sum_{i=1}^{N} T_{si}}{N}
     \label{m1}
\end{equation}
 where  $N$ is the total number of Monte Carlo runs and $T_{si}$ is the source seeking time for the $i^{th}$ run.
\item \textbf{Average number of iterations ($\mu(I)$)}: This metric quantifies the sampling efficiency of the algorithm and is defined in (\ref{m2}) as the average number of waypoints generated to locate the source.  
\begin{equation}
     \mu(I) = \frac{\sum_{i=1}^{N} W_{si}}{N}
     \label{m2}
\end{equation}
where $W_{si}$ is the number of waypoints generated for the $i^{th}$ run to locate the source. 

 
 \item \textbf{Average Swarm Distance ($\mu(SD)$)}: This parameter is related to the fuel efficiency when UAVs are deployed for the search and is defined in (\ref{m3}). Swarm distance is defined as the total distance travelled by the UAVs in the swarm, till the source has been located. 
 \begin{equation}
     \mu(SD) = \frac{\sum_{i=1}^{N} \sum_{j=1}^{n} D_{ji}}{N}
     \label{m3}
\end{equation}
where  $n$ is the size of UAV swarm  and $D_{ji}$ is the distance travelled by the $j^{th}$ UAV in the $i^{th}$ simulation run.

\end{itemize}

The performance comparison is done under different operating conditions as given below.\\
a) \textbf{Inter-UAV communication network topology}: The exchange of information between the UAVs depends on the inter-UAV communication network topology. We consider three scenarios viz. i) fully connected, i)  ring and iii) adaptive toplogies. For fully connected topology, any UAV can communicate with any other UAV in the swarm and is the best case scenario. Ring topology is a  restricted case  where each UAV can communicate only to it's two adjacent neighbours. For adaptive topology, each UAV  communicates with randomly chosen UAVs from the swarm for each waypoint computation. More details of these three topologies are available in \cite{zou2015particle}. The other operating conditions considered are 
b) \textbf{different size of the search region},   c) \textbf{varying number of UAVs in the swarm}, d) \textbf{restricted source movement}, and e) \textbf{measurement noise}.

\subsection{Performance comparison of   source-seeking algorithms}
The average is computed for $N=1000$, for a swarm size $n=5$ for a search space    of dimensions 100 $m$ $\times$100 $m$ unless otherwise stated explicitly.

\subsubsection{\textbf{Effect of inter-UAV communication network topology}}:  The results  shown in Table \ref{topapso1} indicate that APSO is performing better than SPSO and ARPSO in all the three metrices for all the three topologies. The fully connected topolgy gives the best performance as expected. The swarm under adaptive topology performs better than the ring topology. Due to its superior performance, for all the remaining experiments the Fully Connected (FC) topology is used for all the three algorithms.

\begin{table}[h]
\caption{Effect of Network Topology}
\label{topapso1}
\begin{center}
\begin{tabular}{|c|c|c|c|c|}
\hline
Algorithm & Topology & $\mu(I)$ & $\mu(T_s)$ ($s$) & $\mu(SD)$ ($m$)\\
\hline
SPSO & Ring & 43.22 & 34.97  & 387.98 \\
ARPSO & Ring & 41.91 & 34.45 &343.24 \\
\textbf{APSO} & Ring & \textbf{27.36} & \textbf{17.31} &\textbf{175.52} \\
\hline
SPSO & FC &  19.44 & 22.94 & 286.34 \\
ARPSO & FC & 15.28  & 19.59  & 217.2 \\
\textbf{APSO} & FC & \textbf{10.81} & \textbf{11.03} & \textbf{116.58}\\
\hline
SPSO & Adaptive & 38.08  & 32.09  & 371.41 \\
ARPSO & Adaptive & 38.24 &31.78  & 320.69  \\
\textbf{APSO} & Adaptive &\textbf{ 22.3} &\textbf{16.12}  &\textbf{167.41 } \\
\hline
\end{tabular}
\end{center}
\end{table}



\subsubsection{\textbf{Different size of search region}}
The dimensions of the search space is varied from  10 $m$ $\times$ 10 $m$ to 100 $m$ $\times$ 100 $m$. Table \ref{sizearea} shows the performance of all three algorithms with UAVs communicating under fully connected topology. The APSO performs the best among the three algorithms. The improvement in the performance of APSO is higher with an increase in the search area when compared to SPSO and ARPSO algorithm.

\begin{table}[h]
\caption{Effect of Search Area Dimensions}
\label{sizearea}
\begin{center}
\begin{tabular}{|c|c|c|c|c|}
\hline
Algorithm & Area  ($m^2$) & $\mu(I)$ & $\mu(T_s)$ ($s$) & $\mu(SD)$ ($m$)\\
\hline
SPSO          &            & 6.92          & 1.29          & 14.77 \\
ARPSO         &10 $m$ $\times$ 10 $m$ & 6.17          & 1.25          & 14.19 \\
\textbf{APSO} &            & \textbf{5.22} & \textbf{0.80}  &\textbf{8.34} \\

\hline
SPSO          &             & 11.41          & 4.63          & 54.65 \\
ARPSO         & 25 $m$ $\times$ 25 $m$ & 8.85          & 3.74          &43.22 \\
\textbf{APSO} &             & \textbf{7.26} & \textbf{2.37} &\textbf{25.42} \\

\hline
SPSO          &            & 15.63          & 10.67          & 131.71 \\
ARPSO         & 50 $m$ $\times$ 50 $m$& 11.55          & 8.67          & 97.60 \\
\textbf{APSO} &            & \textbf{9.01} & \textbf{5.33} &\textbf{56.28} \\

\hline
SPSO          &            & 17.82          & 16.93          & 209.34 \\
ARPSO         & 75 $m$ $\times$ 75 $m$& 13.58          & 14.20          & 155.29 \\
\textbf{APSO} &            & \textbf{10.18} & \textbf{8.21} &\textbf{86.74} \\
\hline

SPSO          &              & 19.44          & 22.94          & 286.34 \\
ARPSO         & 100 $m$ $\times$ 100 $m$& 15.28          & 19.59          &217.2 \\
\textbf{APSO} &              & \textbf{10.81} & \textbf{11.03} &\textbf{116.58} \\
\hline
\end{tabular}
\end{center}
\end{table}

\subsubsection{\textbf{Number of UAVs in a swarm}}
\begin{figure}[h]
\centering
\includegraphics[width = 8cm, height = 6 cm]{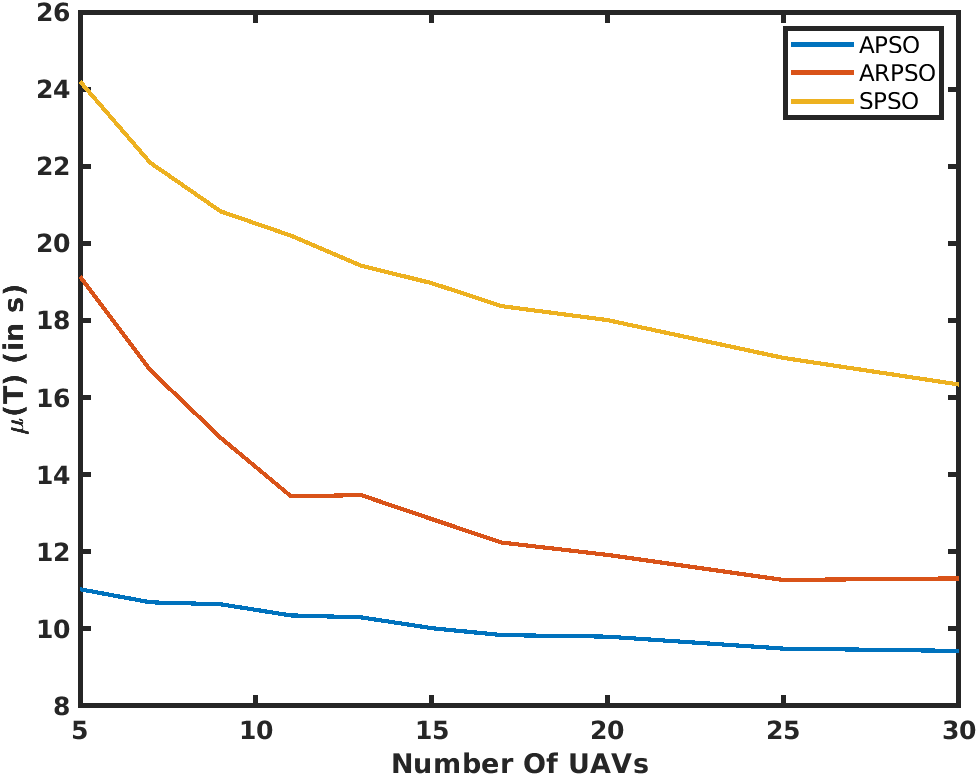}
\caption{Effect of number of UAVs on $\mu(T_s)$}
\label{time1}
\end{figure}

\begin{figure}[h]
\centering
\includegraphics[width = 7cm, height = 6 cm]{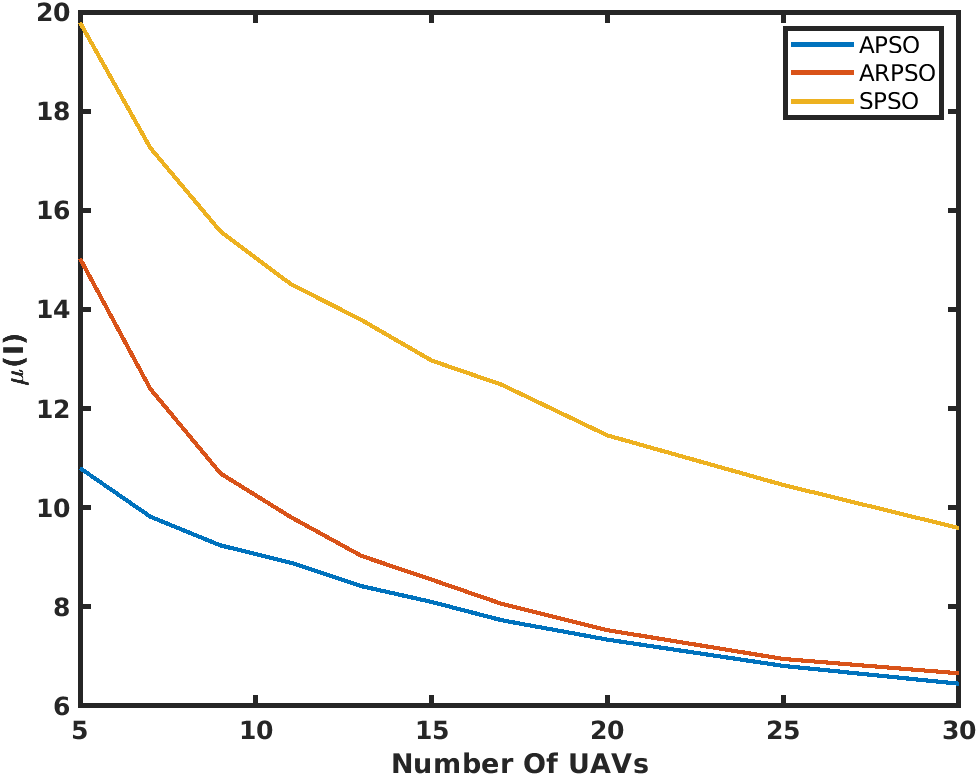}
\caption{Effect of number of UAVs on $\mu$(I)}
\label{Iter1}
\end{figure}

\begin{figure}[h]
\centering
\includegraphics[width = 7cm, height = 6 cm]{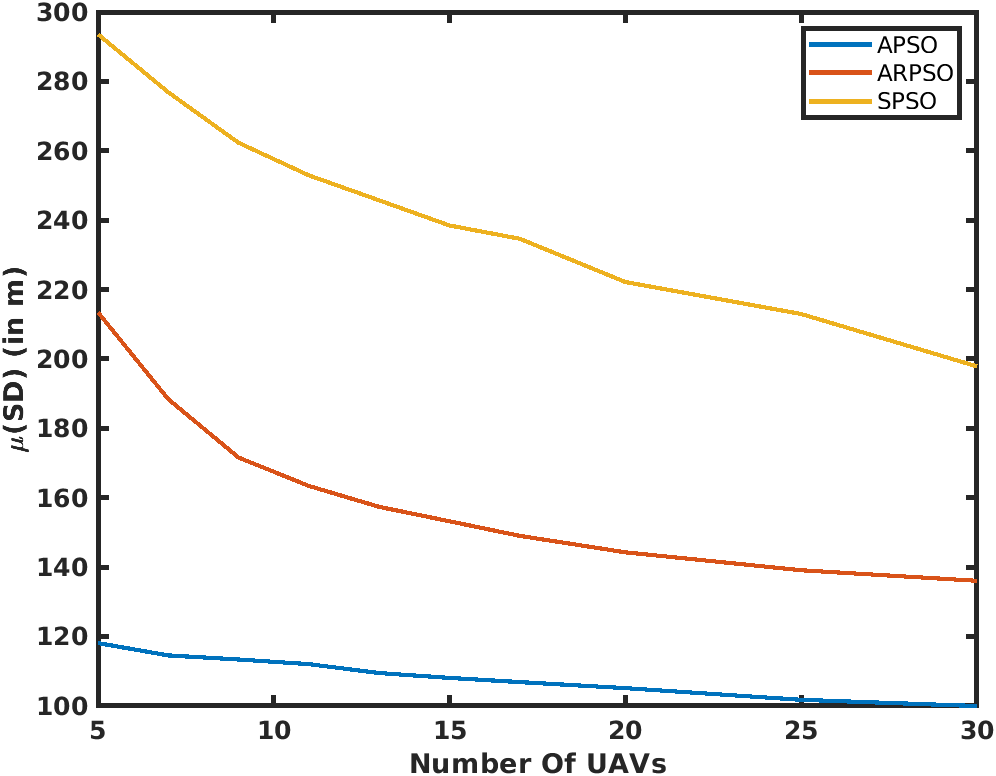}
\caption{Effect of number of UAVs on $\mu$(SD)}
\label{SD1}
\end{figure}

Here the UAV swarm size is increased from 5 till 30 under fully connected topology. The average source seeking time for the three algorithms is plotted in Fig. \ref{time1}. The $\mu(T_s)$ for APSO is almost twice better than SPSO and ARPSO for $n=5$ and later the improvement decreases. The $\mu(T_s)$ saturates after a certain swarm size for all the three algorithms. The comparison for average number of iterations is shown in Fig. \ref{Iter1}. The APSO outperforms SPSO irrespective of the number of UAVs in a swarm. The performance of ARPSO converges to that of APSO for $n\geq20$. The APSO outperforms the other two algorithms for the case of $\mu$(SD) for all the  UAV swarm size as shown in Fig. \ref{SD1}.

\subsubsection{\textbf{Restricted source movement}}
\begin{figure}[h]
\centering
\includegraphics[width = 7cm, height = 5 cm]{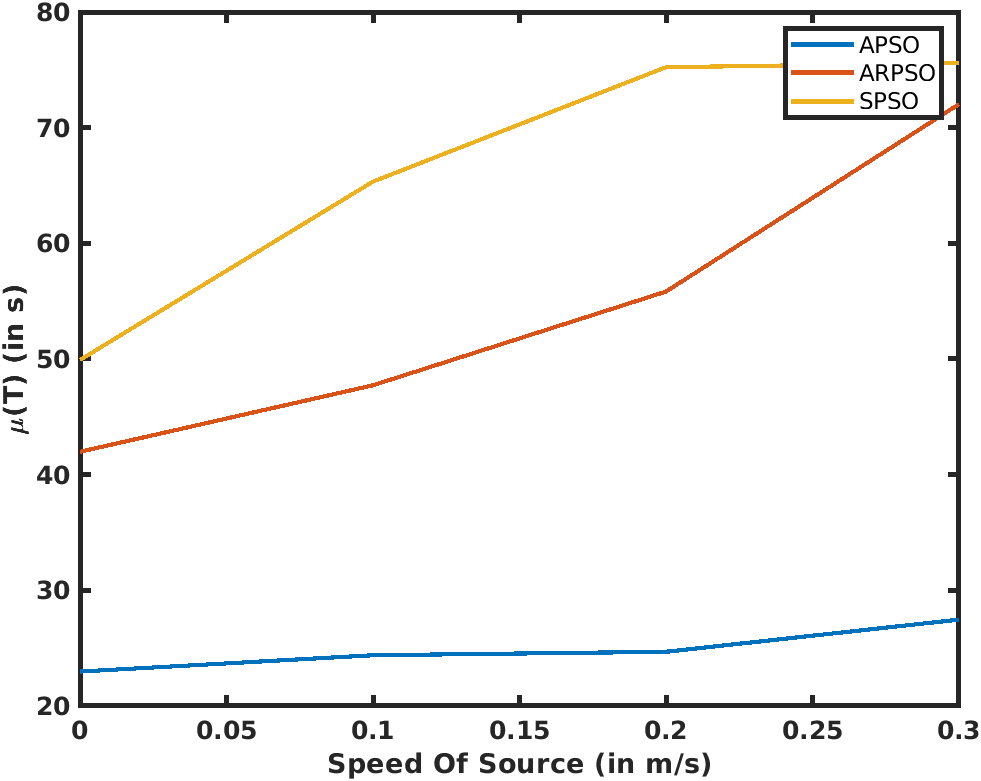}
\caption{Effect of source speed on $\mu(T_s)$}
\label{time2}
\end{figure}

\begin{figure}[h]
\centering
\includegraphics[width = 7cm, height = 6cm]{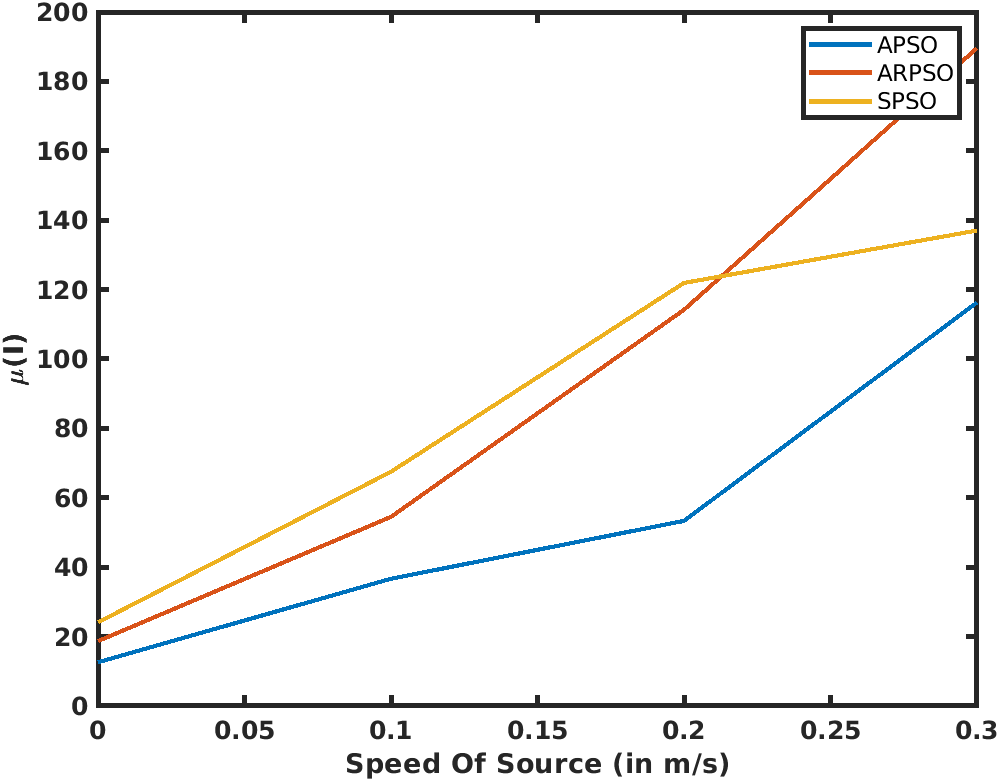}
\caption{Effect of source speed on $\mu$(I)}
\label{AvgIter2}
\end{figure}

\begin{figure}[h]
\centering
\includegraphics[width = 7cm, height = 6 cm]{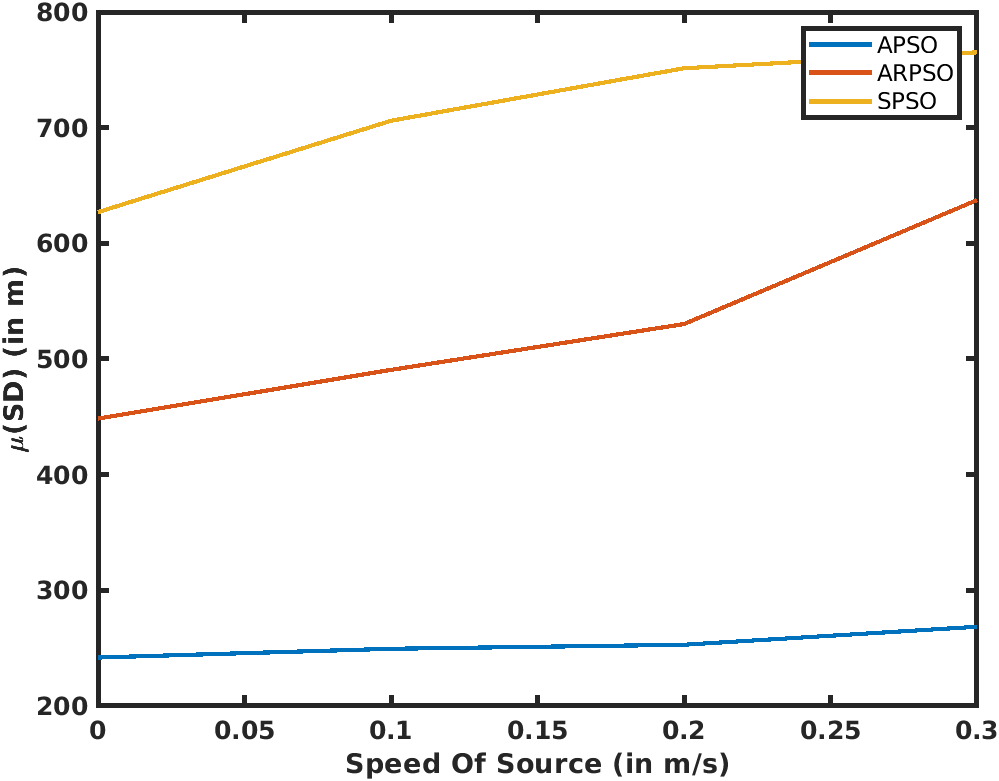}
\caption{Effect of source speed on $\mu$(SD)}
\label{AvgSD2}
\end{figure}

:
In this experiment, the source is  moving randomly with in  a circle of radius  4$m$ and the search space considered here is of 
dimensions 200 $m$ $\times$ 200 $m$. The speed of the source is  varied from 0 to 0.3 $m/s$ and the performance of all the three algorithms is compared. Figure \ref{time2} shows the effect of source speed on the average source seeking time indicating the better performance of APSO when compared to ARPSO and SPSO. The value of $\mu(T_s)$ increases slowly with source speed for APSO,  whereas it increases rapidly for ARPSO and SPSO. Similar result can be observed for the case of $\mu(SD)$ as shown in Fig. \ref{AvgSD2}. The increase in $\mu(I)$ with the source speed is noted for all the algorithms from Fig. \ref{AvgIter2} with APSO performing the best among the three algorithms. For higher source speed, the difference between the performance of APSO and SPSO reduces.

\subsubsection{\textbf{Sensor Noise}}
In real scenarios, the sensor measurement is corrupted by noise as modeled in (\ref{noise1}).
 Gaussian noise with zero mean and standard deviation proportional to the power measured by the UAV is considered as the sensor noise.  The percentage of noise added is increased from 0 to 10  $\%$ to see its effect on the three source seeking algorithms. Additionally, the minimum distance of any UAV  from the source for which the algorithm terminates is increased from 0.1 $m$ to 0.5 $m$ to account for the measurement noise.   The results are tabulated in Table \ref{tnoise}.

\begin{table}[h]
\caption{Effect of Sensor Noise}
\label{topapso}
\begin{center}
\begin{tabular}{|c|c|c|c|c|}
\hline
Algorithm & Sensor Noise ($\%$) & $\mu(I)$ & $\mu(T_s)$ ($s$) & $\mu(SD)$ ($m$)\\
\hline
SPSO          &  & 10.51          & 18.23          & 213.38 \\
ARPSO         & 0& 8.33          & 15.59          &174.80 \\
\textbf{APSO} &  & \textbf{6.94} & \textbf{9.78}  &\textbf{102.63} \\
\hline
SPSO          &  & 22.93          & 25.88          & 278.38 \\
ARPSO         & 0.05& 16.81          & 19.91          &205.31 \\
\textbf{APSO} &  & \textbf{10.66} & \textbf{11.49} &\textbf{114.49} \\
\hline
SPSO          &  & 24.33          & 28.34          & 295.44 \\
ARPSO         & 0.1& 19.06          & 20.04          &201.43 \\
\textbf{APSO} &  & \textbf{12.44} & \textbf{13.17} &\textbf{118.61} \\
\hline
\end{tabular}
\end{center}
\label{tnoise}
\end{table}
The performance of the three algorithms degrades when the Gaussian noise is introduced. However,  APSO is capable of handling the noise much better than the other two algorithms. 
\subsection{DISCUSSION}

  The selection of the parameters $w_1$, $w_2$, $c_1$   and $c_2$ plays a crucial role in the performance of APSO. The parameters $c_1$ and $c_2$ are selected to be the same as the case with  the SPSO algorithm. The stability conditions limits the choice of $w_1$ and $w_2$ to be with in $(-1,\,1)$. The values of  $w_1$ and $w_2$ used here are the ones yielding the best performance. A detailed analysis on the performance of APSO with varying $w_1$, $w_2$, $c_1$ and $c_2$ is not presented here due to the page limit. \\
  The performance of ARPSO algorithm remained closer to the proposed APSO algorithm when compared to the SPSO algorithm. This is mainly due to the adaptive hyperparameters of the ARPSO algorithm when compared to the SPSO algorithm. The improved measurement noise handling capability of  APSO algorithm can be attributed to the underlying third order dynamical system    in contrast to the second order dynamical system for the  ARPSO  and SPSO algorithms.
\section{CONCLUSION AND FUTURE WORK}
This paper presented a new algorithm based on PSO for the multi-UAV source seeking problem. The proposed algorithm implemented a third-order update rule when compared to the existing second-order update rule. The stability  analysis provided a guideline for selecting the hyperparamter values of the algorithm. The convergence of the proposed algorithm is similar to the existing PSO-based source-seeking algorithms [12]. The proposed APSO algorithm showed improved performance when compared to SPSO and ARPSO algorithms under different operating conditions. Obstacle avoidance, dynamic target tracking, and hardware implementation will be interesting extensions of this research work.

\section{ACKNOWLEDGMENT}
The authors thank the financial support provided from the SEED grant of the KCIS project titled "RoboMate: Towards 24/7 Companionship and
Assistance," IIIT Hyderabad to carry out this research work.



\section{References}
[1] Micael S. Couceiro, Rui P. Rocha, and Nuno M. F.
Ferreira. “A novel multi-robot exploration approach
based on Particle Swarm Optimization algorithms”.
In: 2011 IEEE International Symposium on Safety,
Security, and Rescue Robotics. 2011, pp. 327–332.
DOI: 10.1109/SSRR.2011.6106751.

[2] Masoud Dadgar, Shahram Jafari, and Ali Hamzeh. “A
PSO-based multi-robot cooperation method for target
searching in unknown environments”. In: Neurocomputing
177 (2016), pp. 62–74.

[3] Kurt Derr and Milos Manic. “Multi-robot, multi-target
particle swarm optimization search in noisy wireless
environments”. In: 2009 2nd conference on human
system interactions. IEEE. 2009, pp. 81–86.

[4] E. Di Mario, I. Navarro, and A. Martinoli. “A distributed
noise-resistant particle swarm optimization
algorithm for high-dimensional multi-robot learning”.
In: 2015 IEEE international conference on robotics
and automation (ICRA). IEEE. 2015, pp. 5970–5976.

[5] Arjuna Flenner et al. “Levy walks for autonomous
search”. In: Ground/Air Multisensor Interoperability,
Integration, and Networking for Persistent ISR III.
Vol. 8389. International Society for Optics and Photonics.
2012, 83890Z.

[6] K. Harikumar, J. Senthilnath, and S. Suresh. “Multi-
UAV oxyrrhis marina-inspired search and dynamic
formation control for forest firefighting”. In: IEEE
Transactions on Automation Science and Engineering
16.2 (2019), pp. 863–873.

[7] EI Jury. “Robustness of discrete systems: A review”.
In: IFAC Proceedings Volumes 23.8 (1990), pp. 197–
202.

[8] J. Kennedy and R. Eberhart. “Particle swarm optimization”.
In: Proceedings of ICNN’95 - International
Conference on Neural Networks. Vol. 4. 1995, 1942–
1948 vol.4. DOI: 10.1109/ICNN.1995.488968.

[9] Wen-Tsao Pan. “A new Fruit Fly Optimization Algorithm:
Taking the financial distress model as an
example”. In: Knowledge Based Systems - KBS 26
(Feb. 2012). DOI: 10.1016/j.knosys.2011.
07.001.

[10] Eric Rohmer, Surya PN Singh, and Marc Freese.
“V-REP: A versatile and scalable robot simulation
framework”. In: 2013 IEEE/RSJ International Conference
on Intelligent Robots and Systems. IEEE. 2013,
pp. 1321–1326. 

[11] Jason Tillett et al. “Darwinian Particle Swarm Optimization.”
In: Jan. 2005, pp. 1474–1487. 

[12] Ioan Cristian Trelea. “The particle swarm optimization
algorithm: convergence analysis and parameter selection”.
In: Information processing letters 85.6 (2003),
pp. 317–325.

[13] K.A.R. Youssefi and M. Rouhani. “Swarm intelligence
based robotic search in unknown maze-like environments”.
In: Expert Systems with Applications 178
(2021), p. 114907.

[14] Rui Zou et al. “Particle swarm optimization-based
source seeking”. In: IEEE Transactions on Automation
Science and Engineering 12.3 (2015), pp. 865–875.

\end{document}